\documentclass[sigconf,nonacm,natbib=true]{acmart}

\AtBeginDocument{%
  }

\usepackage{times}
\usepackage{latexsym}
\usepackage[T1]{fontenc}
\usepackage[utf8]{inputenc}
\usepackage{microtype}

\usepackage{url}
\usepackage{tabularx}
\usepackage{multirow}
\usepackage{color}
\usepackage{booktabs}
\usepackage{amsmath}

\usepackage{times}
\renewcommand{\footnotesize}{\scriptsize}
\usepackage{graphicx}
\usepackage{enumitem}

\newcommand{\pipeline}{{GINGER}}

\newcommand{\myquestion}[1]{\paragraph*{\textbf{#1}}}

\makeatletter
\renewcommand\paragraph{\@startsection{paragraph}{4}{0pt}%
  {1.5ex \@plus 0.5ex \@minus 0.2ex}%
  {-1em}%
  {\normalfont\normalsize\bfseries\itshape}}
\makeatother

\newcommand{\myparagraph}[1]{\paragraph*{#1}}

\setcopyright{cc}
\copyrightyear{2025}
\acmYear{2025}
\acmDOI{XXXXXXX.XXXXXXX}
\acmConference[SIGIR '25]{The 48th International ACM SIGIR Conference on Research and Development in Information Retrieval}{July 13--18, 2025}{Padua, Italy}
\acmISBN{978-1-4503-XXXX-X/2025/07}

\begin{document}

\title{GINGER: Grounded Information Nugget-Based Generation of Responses}

\author{Weronika Łajewska}
\orcid{0000-0003-2765-2394}
\affiliation{%
  \institution{University of Stavanger}
  \city{Stavanger}
  \country{Norway}
}
\email{weronika.lajewska@uis.no}

\author{Krisztian Balog}
\orcid{https://orcid.org/0000-0003-2762-721X}
\affiliation{%
  \institution{University of Stavanger}
  \city{Stavanger}
  \country{Norway}
}
\email{krisztian.balog@uis.no}

\begin{abstract}
Retrieval-augmented generation (RAG) faces challenges related to factual correctness, source attribution, and response completeness. To address them, we propose a modular pipeline for grounded response generation that operates on information nuggets---minimal, atomic units of relevant information extracted from retrieved documents. The multistage pipeline encompasses nugget detection, clustering, ranking, top cluster summarization, and fluency enhancement. 
It guarantees grounding in specific facts, facilitates source attribution, and ensures maximum information inclusion within length constraints. 
Extensive experiments on the TREC RAG'24 dataset evaluated with the AutoNuggetizer framework demonstrate that \pipeline{} achieves state-of-the-art performance on this benchmark.
\end{abstract}
 

\maketitle

\section{Introduction}

We observe an increasing reliance on conversational assistants, such as ChatGPT, for a variety of information needs, involving open-ended questions, indirect answers that require inference, and complex queries with partial answers spread over several passages~\citep{Bolotova-Baranova:2023:ACL, Zamani:2023:FNT, Gabburo:2024:arXiv}.
Generating responses poses challenges related to factual correctness~\citep{Ji:2023:ACMa, Koopman:2023:EMNLP, Tang:2023:ACL}, source attribution~\citep{Rashkin:2021:Comput.}, information verifiability~\citep{Liu:2023:EMNLP}, consistency, and coverage~\citep{Gienapp:2024:SIGIR}. Recently proposed retrieval-augmented generation (RAG) models aim to address these issues by conditioning the generative processes on retrieved sources, yet ensuring response transparency and source attribution remains an open challenge~\citep{Lewis:2020:NIPS, Huang:2024:arXiv, Gienapp:2024:SIGIR}.
Current commercial generative search engines appear informative but often contain unsupported statements and inaccurate citations, further highlighting the difficulty of grounding responses~\citep{Liu:2023:EMNLP}.
Injecting evidence in LLM prompts can reduce hallucinations~\citep{Koopman:2023:EMNLP}. However, redundant information and overly long contexts can lead to the ``lost in the middle'' problem, where models struggle to retrieve relevant information from the middle of long contexts~\citep{Liu:2024:Trans.}. 
Consequently, a post-retrieval refinement step is recommended to retain only essential information and prevent key details from being diluted~\citep{Gao:2023:arXiv}.

To address the above challenges, we present a modular pipeline for \textbf{G}rounded \textbf{I}nformation \textbf{N}ugget-Based \textbf{GE}neration of
\textbf{R}esponses (\pipeline{}). The main novelty of our approach compared to existing RAG approaches~\citep{Ram:2023:Trans., Shi:2023:arXiv} is that it operates on \emph{information nuggets}, which are atomic units of relevant information~\citep{Pavlu:2012:WSDM}.
Given a set of passages retrieved in response to a user query, our approach identifies information nuggets in top passages, clusters them by query facet, ranks clusters by relevance, summarizes the top ones, and refines the response for fluency and coherence.
\pipeline{} uniquely models query facets to ensure the inclusion of the maximum number of unique pieces of information answering the question.
This approach can significantly improve the user experience by ensuring the grounding of the final response in the source passages and enabling easy verifiability of source attribution. 

We evaluate our system on the augmented generation task within the recently launched Retrieval-Augmented Generation track at Text REtrieval Conference (TREC RAG'24). 
We report on official evaluation results based on AutoNuggatizer~\citep{Pradeep:2024:arXiv} evaluation framework piloted in this track for variants of our pipeline submitted to TREC RAG'24. For other system configurations, we utilize a reimplementation of this evaluation framework (using the same prompts) to score responses.
The results demonstrate that \pipeline{} achieves performance on par with top-performing TREC submissions and also significantly outperforms two strong baselines: (1) a grounded generation approach for open-domain QA~\citep{Ren:2023:arXiv} and (2) chain-of-though generation with in-context learning demonstration~\citep{Wei:2022:NIPS}.
We show that \pipeline{} successfully filters out and synthesizes information from relevant sources, and its performance increases with the amount of information provided to the system.
Finally, an ablation study reveals that the core strength of \pipeline{} lies in its operation on information nuggets, rather than the individual performance of its components.

In summary, the main contributions of this work are twofold: 
(1) A novel modular response generation pipeline that operates on information nuggets, enabling precise grounding and facilitating source verification. 
(2) Extensive experimental evaluation of the proposed approach on the TREC RAG'24 augmented generation task using the AutoNuggetizer framework.
All the resources developed in this paper, along with additional results and analysis are available online: \url{https://github.com/iai-group/ginger-response-generation}.

\begin{figure*}[tp]
    \centering
    \includegraphics[width=0.85\textwidth]{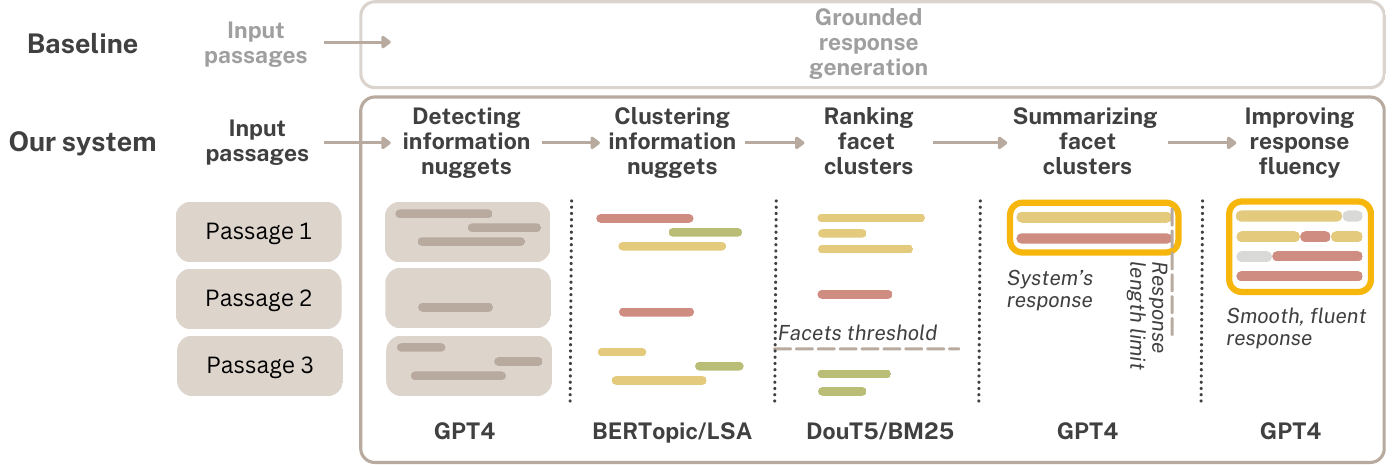}
    \caption{High-level overview of our nugget-based response generation pipeline (\pipeline{}).}
    \label{fig:system_schema}
\end{figure*}
\section{Related Work}

\myparagraph{Retrieval-Augmented Generation}
Unlike traditional search engines that return a ranked list of documents, RAG systems generate a single, coherent response by synthesizing diverse perspectives from retrieved sources, conditioning the generation process on retrieved evidence either through prompt injection or direct attention during inference~\citep{Gienapp:2024:SIGIR, Mialon:2023:arXiv, Izacard:2021:EACL, Shi:2023:arXiv, Ram:2023:Trans.}.
RAG approaches include conditioning models on retrieved document chunks~\citep{Borgeaud:2022:ICML} or combining parametric and non-parametric memory~\citep{Lewis:2020:NIPS, Huang:2024:arXiv}. 
In RAG, retrieving documents based on partial relevance can introduce irrelevant or loosely related text, which may hinder the quality of the generated response~\citep{Cuconasu:2024:arXiv}.
Also, generation performance can significantly drop when relevant information is not at the beginning or end of long contexts~\citep{Liu:2024:Trans.}. 
These findings suggest that basic retrieve-then-generate pipelines may not effectively reduce hallucinations~\citep{Koopman:2023:EMNLP}. 
Several methods have been proposed for context curation for LLMs, such as using small language models to remove unimportant tokens~\citep{Jiang:2023:arXiv}, 
information aggregation using multi-agent collaboration~\citep{Zhang:2024:arXiv}, or
training information extractors and condensers using contrastive learning techniques~\citep{Xu:2023:arXivb, Yang:2023:EMNLP}.
Generative responses often exhibit high fluency and perceived utility but frequently contain unsupported statements or inaccurate citations~\citep{Liu:2023:EMNLP}. 
Despite advancements in LLMs, abstractive summaries still suffer from hallucinations and factual errors~\citep{Ladhak:2022:ACL, Tang:2023:ACL, Falke:2019:ACL, Tang:2022:NAACL-HLT, Ji:2023:ACMa, Koopman:2023:EMNLP}. Source attribution, which measures the accuracy and support of generated statements through citations~\citep{Rashkin:2021:Comput.}, and verifiability, which necessitates each statement to be fully supported by in-line citations, are key requirements to address these challenges~\citep{Liu:2023:EMNLP, Schuster:2023:NAACL-HLT}.
To ensure high citation precision while maintaining fluency, we propose to extract and group atomic statements (referred to as ``atomic/semantic content units''~\citep{Nenkova:2007:ACM, Liu:2023:ACL} or ``information nugget'' in traditional IR~\citep{Pavlu:2012:WSDM, Sakai:2023:arXiv}) from sources and summarize them with LLMs.

\myparagraph{TREC RAG'24}
The RAG track at TREC has been launched in 2024 with a focus on combining retrieval methods for finding relevant information within large corpora with LLMs to enhance the ability of systems to produce relevant, accurate, and contextually appropriate content~\citep{Pradeep:2024:arXiv}. The track is divided into three tasks: Retrieval (R), which involves ranking and retrieving the most relevant segments from the corpus; Augmented Generation (AG), which requires generating RAG answers using top-k relevant segments from a baseline retrieval system provided by organizers; and Retrieval-Augmented Generation (RAG), where participants generate RAG answers with attributions using their retrieval system and chunking technique. Our work focuses on the augmented generation task, similar to in-context retrieval augmented language models without modifying model weights~\citep{Ram:2023:Trans., Muhlgay:2023:EACL}.

A trend toward retrieval context curation is evident in TREC RAG'24 submissions. 
A common approach to AG task involves generating responses based on the top 20 retrieved documents, often in a single step using proprietary models, with an optional post-processing phase. Several submissions adopt a multi-step approach, such as segment clustering followed by 
extracting, combining, and condensing relevant information. 
Similarly, some approaches emphasize verifying key facts across multiple documents, eliminating redundant content, prioritizing facts by relevance, and enhancing clarity and coherence.
In this work, we take this a step further by not only curating the LLM context but also decomposing the response generation process to mitigate the negative effects of irrelevant and redundant information by operating on atomic pieces of information.
\section{Nugget-Based Multi-Step Response Generation}

We present a novel method for generating grounded conversational responses by operating on information nuggets. 
It explicitly models various facets of the query based on the retrieved information and generates a concise response that adheres to length constraints. 

Generating grounded responses is a multistage process, illustrated in Fig.~\ref{fig:system_schema}, that includes: (1) detecting information nuggets in top relevant passages, (2) clustering detected nuggets, corresponding to different facets of the query, (3) ranking the clusters with respect to their relevance to the query, (4) summarizing the top-ranked clusters to be included in a final response, and (5) refining the response to improve its fluency and coherence.
Steps 1-3 aim at curating the context for response generation, while Steps 4-5 focus on synthesizing and refining the response.
By operating on information nuggets in all intermediate components of the pipeline we ensure the grounding of the final response in the source passages, ensuring that all information in the final response is entailed by the source~\citep{Falke:2019:ACL}. 
The prompts used by different components of the system are made available in the online appendix: \url{https://bit.ly/40ZkFJ8}.

\myparagraph{Detecting Information Nuggets}
We automatically detect information nuggets by prompting an LLM to annotate input passages with information nuggets containing the key information that answers the query. Specifically, it is instructed to copy the passage's text and place the annotated information nuggets between specific tags, without modifying the passage content or adding any extra symbols. 

\myparagraph{Clustering Information Nuggets}
Next, we cluster the detected information nuggets with respect to different facets of the query topic. 
This clustering step serves two purposes.
First, it addresses the problem of information redundancy, stemming from the fact that information nuggets and their variants can appear in multiple documents in different forms but still convey the same information~\citep{Pavlu:2012:WSDM}.
Second, by clustering redundant information nuggets, we attempt to increase the information density of the generated information~\citep{Adams:2023:ACL}.
Nugget clustering is challenging due to the semantic closeness of nuggets within the same topic.
We address this by employing a neural topic modeling technique, BERTopic~\citep{Grootendorst:2022:arXiv}, and adjusting its sensitivity on the validation partition of the dataset.
Ideally, information nuggets in each cluster represent specific facets of the answer to the query.

\myparagraph{Ranking Facet Clusters}
This step in the pipeline is responsible for the ranking of facet clusters with respect to the input query to determine which clusters are most important and should be prioritized for inclusion in the response, and which may be skipped~\citep{Gao:2023:arXiv, Liu:2024:Trans.}. Given the relatively low number of facet clusters we observe in practice, we can employ more expensive reranking techniques relying on pairwise comparisons to maximize effectiveness. 
Specifically, we employ pairwise reranking using  duoT5~\citep{Pradeep:2021:arXiv} by joining the nuggets in clusters and treating them as individual passages.

\myparagraph{Summarizing Facet Clusters}
The response is made up of the summaries of the top $n$ clusters, where $n$ is the facet threshold that controls the desired response length and may be adjusted based on the information need, task context, or user preferences. 
Each cluster of information nuggets is summarized independently as a single sentence with the maximum number of words specified in the prompt, to stop the LLM from generating very long sentences~\citep{Goyal:2023:arXiv}. We follow the prompt design used for short story summarization~\citep{Subbiah:2024:arXiv} to generate summaries that are short, concise, and only contain the information provided.
Previous steps in the pipeline ensure that the most relevant information from retrieved passages is synthesized and that the generated summaries are attributed to the sources. This allows summarization to operate in a shorter but more relevant context.

\myparagraph{Improving Response Fluency}
Our modular approach results in a response that is a concatenation of independent summaries of facet clusters, that may lack fluency and consistency. To mitigate this shortcoming, we include an additional step to rephrase the generated response with the help of an LLM. The LLM is prompted not to modify the provided information, nor include any additional content.
\begin{table}[tp]
    \caption{Response evaluation with AutoNuggetizer. TREC scores are provided for TREC RAG'24 AG submissions. The remaining scores are based on our reimplementation of the framework.}
    \label{tab:eval_pipeline_rag}
    \centering
    \footnotesize
    \begin{tabular}{lccccc}
        \toprule
        \multirow{2}{*}{\textbf{Method}} &\multicolumn{5}{c}{\textbf{V\_strict}} \\
        \cmidrule{2-6}
        &\textbf{TREC} &\textbf{GPT4o} &\textbf{Gemini} &\textbf{Claude} &\textbf{avg LLM} \\
        \midrule
        baseline-top5 &0.247 &0.332 &0.468 &0.525 &0.442 \\ 
        baseline\_CoT-top5 & --- & 0.332 & 0.452 & 0.500 & 0.428 \\
        \midrule
        Webis & 0.357 &  --- & ---  & ---  & ---  \\
        TREMA & 0.261 &  --- & ---  & ---  & ---  \\
        \midrule
        \pipeline{}-top20 wo/ rewriting &\textbf{0.427} &\textbf{0.500} &\textbf{0.543} &\textbf{0.659} &\textbf{0.568} \\
        \pipeline{}-top10 wo/ rewriting &0.369 &0.423 &0.502 &0.582 &0.502 \\ 
        \pipeline{}-top5 wo/ rewriting &0.213 &0.263 &0.392 &0.431 &0.362 \\
        \pipeline{}-top5  &0.211 &0.279 &0.400 &0.451 &0.377 \\
         \bottomrule
    \end{tabular}
\end{table}

\begin{table}[tp]
    \caption{Response evaluation with AutoNuggetizer of responses generated with different variants of \pipeline{} without the fluency enhancement step and with top 20 passages provided as input.}
    \label{tab:eval_pipeline_variants}
    \centering
    \small
    \begin{tabular}{llc}
        \toprule
        \textbf{Clustering} & \textbf{Ranking} & \textbf{V\_strict (avg LLM)} \\
        \midrule
        BERTopic & DuoT5 & \textbf{0.568} \\
        BERTopic & BM25 & 0.554 \\
        LSA & DuoT5 & 0.521 \\
        LSA & BM25 & 0.551 \\
         \bottomrule
    \end{tabular}
\end{table}

\section{Experimental Setup}

This section presents the evaluation dataset, baselines, and implementation details. 

\myparagraph{Dataset}

We base our evaluation on the Augmented Generation task of TREC RAG'24~\citep{Pradeep:2024:arXiv} comprising 301 queries.  
This work focuses on the problem of response generation and assumes a ranked list of passages to be provided, along with the query, as input. 
In all experiments, we utilize the top relevant passages (from MS MARCO V2.1 segment collection) from a fixed list of 100 retrieved results provided by the organizers. This setup represents the real-world case with the top passages retrieved by a competitive passage ranker. 
To account for the amount of input information, we consider three sizes of input rankings containing 5, 10, or 20 passages.

\myparagraph{TREC Evaluation}

We use the AutoNuggetizer framework proposed for RAG evaluation and validated during  TREC RAG'24~\citep{Pradeep:2024:arXiv}. AutoNuggetizer comprises two steps: nugget creation and nugget assignment. In nugget creation, nuggets are formulated based on relevant documents and classified as either ``vital'' or ``okay''~\citep{Voorhees:2004:NIST}. The second step, nugget assignment, involves assessing whether a system’s response contains specific nuggets from the answer key. 
The score \(V_{strict}\) for system's response is defined as follows:
\[V_{strict}=\frac{\sum_{i}{ss_i^v}}{|n^v|}\] where \(n^v\) 
represents the subset of the vital nuggets; \(ss^v_i\) is 1 if the response supports the \emph{i}-th nugget and is 0 otherwise.
The score of a system is the mean of the scores across all queries.

\myparagraph{LLM-based Evaluation}
We reimplemented the AutoNuggetizer evaluation framework to compute the $V_{strict}$ measure, adhering to the original prompts from~\citet{Pradeep:2024:arXiv}.
To make the evaluation more robust and mitigate any potential bias of using the same LLM for response generation and judging, we use the average of the scores generated by three different LLMs (\emph{gpt-4o-2024-08-06}, \emph{claude-3-5-haiku-20241022}, \emph{gemini-1.5-flash}) as a final score (as opposed to original TREC RAG scores which are based solely on GPT-4o). We validate our implementation of the evaluation framework by comparing the results for our submitted runs with the official numbers reported by track organizers~\citep{Pradeep:2024:arXiv}. Even though our V\_strict scores are higher than the scores reported in the TREC RAG track, the relative ordering of the systems remains the same.

\myparagraph{Baseline}

To compare our method with models using external knowledge in generation, we focus on grounded response generation with a fixed retriever. This excludes standard end-to-end RAG models~\citep{Lewis:2020:NIPS, Guu:2020:ICML, Izacard:2023:JMLR} and LLM approaches relying solely on internal knowledge~\citep{Sun:2022:ICLR}. 
In pursuit of a strong baseline, we explored grounded text generation, Chain-of-Thought (CoT) prompting, and open-domain QA models using samples from the TREC CAsT'22 datasets~\citep{Owoicho:2022:TRECb}.
We found that a pre-trained model proposed for grounded text generation in dialogues that rely on external knowledge~\citep{Peng:2022:arXiv} tended to copy directly from the source text, showing extractive behavior.
Using an off-the-shelf LLM proposed for a retrieval-augmented setting for open-domain QA showed notably better coherence and naturalness without additional training~\citep{Ren:2023:arXiv, Ram:2023:Trans., Muhlgay:2023:EACL}.
Therefore, our first baseline is OpenAI's GPT-4 model, prompted similarly to retrieval-augmented QA~\citep{Ren:2023:arXiv} using the top 5 relevant passages (\emph{baseline-top5}).
The length of the generated summary is limited to around 100 words (3 sentences), controlled via the task model prompt~\citep{Goyal:2023:arXiv}. 
The second baseline uses GPT-4 with Chain-of-Thought prompting~\citep{Wei:2022:NIPS} and one ICL demonstration created manually based on TREC CAsT'22 dataset~\citep{Owoicho:2022:TRECb} (\emph{baseline\_CoT-top5}).

\myparagraph{Implementation Details}
\label{sec:implementation}

The modular design of \pipeline{} allows for independent implementation of individual components of the pipeline. 
The detection of information nuggets is performed with GPT-4, with the query and passage as input.
Information nuggets clustering is based on BERTopic, with parameters set experimentally on samples from TREC CaST'22.\footnote{If fewer than four information nuggets are identified in the top $n$ passages, we skip clustering; instead, we treat each nugget as an independent cluster and proceed with standard ranking and summarization.} 
The ranking of information nugget clusters is done using a duoT5 model,\footnote{\url{https://huggingface.co/castorini/duot5-base-msmarco}} implemented based on the HuggingFace transformers library.
The top 3 ranked information nuggets clusters are passed to query biased summarization~\citep{Tombros:1998:SIGIR} performed with GPT-4. The length of each cluster summary is limited to one sentence and around 35 words and specified in the prompt~\citep{Goyal:2023:arXiv}; the output length limit is specified in the model parameters.
\section{Results and Analysis}

We seek to measure \pipeline{}'s capability to generate high-quality responses.
Table~\ref{tab:eval_pipeline_rag} presents results for the two baseline systems, as well as for our nugget-based response generation pipeline (with or without response fluency improvement) using different numbers of retrieved passages. Responses from the top 5 passages are limited to 3 sentences, while those from the top 10 or 20 passages have a 400-word limit. TREC scores are reported for system configurations where we have official evaluation results. 

\myquestion{Does \pipeline{} improve response generation performance over the baselines?}

The best-performing variant of our system outperforms both baseline approaches. Even prompting the model to divide the task into several steps using Chain-of-Thought and providing a ground-truth response as an example does not help in the response generation process. This implies that given the complexity of the queries and the amount of input context to be taken into account, the LLM needs further guidance to generate an accurate answer.

\myquestion{How does \pipeline{} perform in comparison to other systems submitted to TREC RAG'24?}

Based on the initial results provided by TREC RAG organizers~\citep{Pradeep:2024:arXiv}, our best system (\emph{\pipeline{}-top20 wo/ rewriting}) is among the top performing AG submissions. Even though several other systems decompose response generation into multi-step process, \pipeline{} shows higher performance. For reference, we include two other competitive AG submissions in the results table (\emph{Webis.webis\--ag-run0-taskrag} and \emph{TREMA-UNH.Enhanced \_Iterative\_Fact\-\_Refinement\_and\_Prioritization}). 

\myquestion{How does the amount of input information affect \pipeline{}'s performance?}

By operating on information nuggets throughout the pipeline, \pipeline{} grounds responses in specific facts and synthesizes information from the provided passages. Responses generated from more passages are of higher quality, indicating that the additional context is indeed utilized in the system output (\emph{\pipeline{}-top10 wo/ rewriting} vs. \emph{\pipeline{}-top20 wo/ rewriting}). This suggests that splitting response generation into several independent steps and increasing the granularity of information mitigates the information loss to which LLMs are prone. Even with potential information redundancy, responses of the same length limit score higher with more input context, implying that they include more vital facts.

\myquestion{How much do individual pipeline components contribute to overall system performance?}

\pipeline{}'s modular architecture allows us to evaluate individual components and their impact on the final responses.  We experiment with traditional Latent Semantic Analysis (LSA) for nugget clustering and BM25 for cluster ranking, as alternatives.  Following our best-performing setup, we provide 20 relevant passages as input and limit responses to 400 words.
Table~\ref{tab:eval_pipeline_variants} presents the results of this ablation study.
We find that modifying specific components does not strongly impact response quality. Therefore, we conclude that operating on information nuggets, rather than the effectiveness of individual components, is the primary factor contributing to \pipeline{}'s performance. This highlights the importance of higher information granularity and reduced redundancy in response generation.

\myquestion{Does LLM-based fluency enhancement reduce grounding?}

Since the responses returned by our method are concatenations of independent facet cluster summaries, the final response may lack fluency and coherence. However, the difference between the responses generated with the fluency improvement and without this step (\emph{\pipeline{}-top5} vs \emph{\pipeline{}-top5 wo/ rewriting}) is not significant. This indicates that LLMs can be used to refine response fluency without sacrificing quality or grounding.
\section{Conclusions}

We have introduced an approach for generating grounded responses that utilizes information nuggets from the top retrieved passages and employs a multi-stage process (clustering, reranking, summarization, fluency enhancement) to generate concise, information-rich text that is free of redundancy.
Our approach offers several key advantages: maximizing information within response length limits and ensuring grounding in input context.
Automatic evaluation with AutoNuggetizer shows that \pipeline{} outperforms two strong baselines and performs comparably to top TREC RAG'24 submissions. 
Our modular method lays a strong foundation for future research on constraint-based response generation. This could include investigating the impact of various response lengths (e.g., based on user preferences), controlling response completeness, or developing strategies to manage redundancy in multi-turn conversations by considering previously discussed facets.

\section*{Acknowledgmens}
This research was supported by the Norwegian Research Center for AI Innovation, \grantsponsor{NorwAI}{NorwAI}{https://www.ntnu.edu/norwai} (Research Council of Norway, nr.~\grantnum{NorwAI}{309834}).

\bibliographystyle{ACM-Reference-Format}
\bibliography{ecir2025-response_completeness}

\end{document}